%% file: template.tex
\documentclass{article}
\usepackage{arxiv}
\usepackage[utf8]{inputenc} 
\usepackage[T1]{fontenc}    
\usepackage{hyperref}       
\usepackage{url}            
\usepackage{booktabs}       
\usepackage{amsfonts}       
\usepackage{nicefrac}       
\usepackage{microtype}      
\usepackage{graphicx}
\usepackage{natbib}
\usepackage{doi}
\usepackage{tablefootnote}
\usepackage{float}
\usepackage{amsmath}
\usepackage{subcaption}
\usepackage{tikz}
\usetikzlibrary[positioning]
\usetikzlibrary{patterns}
\title{Detecting Fetal Alcohol Spectrum Disorder in children using Artificial Neural Network}

\author{ \href{https://orcid.org/0000-0001-5399-6620}{\includegraphics[scale=0.06]{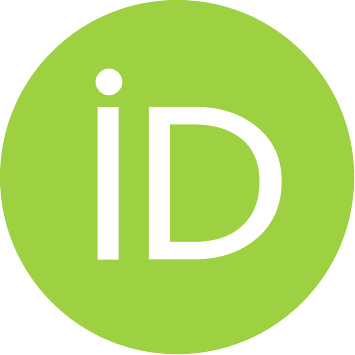}\hspace{1mm}Vannessa de J Duarte C} \\
	Escuela de Ciencias Empresariales\\
	Universidad Catolica del Norte\\
	Coquimbo  \\
	\And
	\href{https://orcid.org/0000-0003-0969-5139}{\includegraphics[scale=0.06]{orcid.pdf}\hspace{1mm}Paul Leger} \\
	Escuela de Ingenieria\\
	Universidad Catolica del Norte\\
	Coquimbo, Chile  \\
	\And
	\href{https://orcid.org/0000-0003-3999-1916}{\includegraphics[scale=0.06]{orcid.pdf}\hspace{1mm}Sergio A. Contreras} \\
	Escuela de Ciencias Empresariales\\
	Universidad Catolica del Norte\\
	Coquimbo, Chile  \\
	\texttt{scontreras@ucn.cl} \\
	\And
	\href{https://orcid.org/0000-0003-1228-3186}{\includegraphics[scale=0.06]{orcid.pdf}\hspace{1mm}Hiroaki Fukuda} \\
	Shibaura Institute of Technology\\
	Tokio, Japan
}

\renewcommand{\headeright}{Technical Report}
\renewcommand{\undertitle}{Technical Report}
\renewcommand{\shorttitle}{FASD}

\hypersetup{
pdftitle={A template for the arxiv style},
pdfsubject={q-bio.NC, q-bio.QM},
pdfauthor={David S.~Hippocampus, Elias D.~Striatum},
pdfkeywords={First keyword, Second keyword, More},
}

\begin{document}
\maketitle

\begin{abstract}
     Fetal alcohol spectrum disorder (FASD) is a syndrome whose only difference compared to other children's conditions is the mother's alcohol consumption during pregnancy. An earlier diagnosis of FASD improving the quality of life of children and adolescents. For this reason, this study focus on evaluating the use of the artificial neural network (ANN) to classify children with FASD and explore how accurate it is. ANN has been used to diagnose cancer, diabetes, and other diseases in the medical area, being a tool that presents good results. The data used is from a battery of tests from children for 5-18 years old (include tests of psychometric, saccade eye movement, and diffusion tensor imaging (DTI)). We study the different configurations of ANN with dense layers. The first one predicts 75\% of the outcome correctly for psychometric data. The others models include a feature layer, and we used it to predict FASD using every test individually. The models accurately predict over 70\% of the cases, and psychometric and memory guides predict over 88\% accuracy. The results suggest that the ANN approach is a competitive and efficient methodology to detect FASD. However, we could be careful in used as a diagnostic technique.
\end{abstract}
\section{Introduction}

The consumption of alcohol during pregnancy is an indisputable generator of mental health problems in Children \citep{o2012guidelines, mccallum2018drink}.  Conditions that emerge from this consumption are grouped under the concept of fetal alcohol spectrum disorder (FASD). This concept was coined in Canada, and included effects may include physical, mental, behavioral, and learning disabilities with possible lifelong implications diagnosis. The most common conditions are fetal alcohol syndrome (FAS), fetal partial alcohol syndrome (pFAS), and the Alcohol-related neurodevelopmental disorder (ARND) \citep{sun_revisiting_2017,cook_fetal_2016}, being the FAS is the most severe consequence of alcohol consumption since the infants would have potential mental retardation. Most of the common outcomes include significant hyperactivity, language delay, and school learning problems. There is an observed higher rate of mental health problems in adolescence, problematic use of alcohol and other drugs.

An earlier diagnosis of these conditions could reduce the long-term health and psychosocial outcome. Especially cognitive difficulties. Thus, improving the quality of life of children and adolescents . In addition, people who are been diagnosed with FASD report higher unemployment, law-breaking, and other social behavior problems. These problems result in a higher cost to the public welfare and society at large.
In spite of the importance of the diagnosis of FASD, currently, there are no direct clinical characteristics that can be used to diagnose a child with FASD. The only distinct difference compared to other childhood or adolescent conditions, is the mother’s alcohol consumption during pregnancy \citep{chudley_diagnosis_2018, popova2018world, popova_estimation_2017}. If there is no certainty about the mother’s alcohol consumption. some indirect effects are associated with FAS, such as facial dysmorphia, growth deficiency, or central nervous system abnormalities. That can be used as proxy for the diagnosis. However, when none of these effects are present, the diagnosis becomes challenging. \citep{paolozza_diffusion_2016, paolozza_altered_2013}.

For the previous reasons, an increased effort to develop techniques based on specific behavior, profiles to allows an early diagnose of FASD. Yet these tools are failing to make a specific diagnosis of FASD compared to other neurodevelopmental conditions \cite{may_prevalence_2018, popova2018world}.  Among these tools, Machine Learning (ML) has become one of the prime tools to improve FASD detection \cite{rodriguez2021detection}. ML are computational algorithms designed to emulate human intelligence at learning from the surrounding environment \cite{el2015machine}. The areas of application in medicine are varied. Such as optical character recognition, face recognition, scientific image analysis, biological signal analysis, and others. In the case of diagnosis, machine learning techniques that have been used for medical diagnosis are: Pattern classification, \citep{fang_facial_2009}, Principal Component Analysis (PCA) \citep{huang_using_2005}, Vector Support Machines (SVM) \cite{kim_pre-operative_2011}, Deep Learning \citep{li_deep_2019, ravi_deep_2017, sun_revisiting_2017},and, especially Artificial Neural Networks (ANN) \citep{al2011artificial,lundervold_overview_2019, wozniak_what_2011}. The latter has become popular for its model estimation transparency and have been used to help diagnose diseases such as cancer and even FASD from brain volume images \cite{little_multivariate_2020}.nd have been used to help diagnose diseases. Such as cancer and even FASD from brain volume image \citep{little_multivariate_2020}.

In consequence, this paper will try to fill the gap of the use of ML techniques for an improved FASD diagnosis. Specifically, this work is focuses on developing computational algorithms based on ANN to classify children with FASD. The research question is whether an ANN model could be used for medical diagnosis. One of the importance of testing ANN is the easy implementation and simplicity in applying existing data. This model will be evaluated based on its performance. In terms of precision, completeness, and performance comparison with other machine learning techniques. Used for differentiation using various tests such as psychometric, saccadic eye movement, and diffusion tensor imaging (DTI). It is believed that techniques based on learning machines are a good tool. For grouping and comparing the clinical information obtained from a set of psychometric data. Thus expanding the possibility of detecting differences in FASD patients. To our knowledge, the use of machine learning as FASD diagnostic tools has not been tested. However there are current studies that have been used them such as \cite{zhang_detection_2019}. However they have used support vector machine regression (SVMR) algorithms to analyze psychometric data and DTI and other methods for the remaining test. 
This paper is organized as follows: the next section presents materials and methods are used to carry out this study. Additionally describing the experiment execution. Section 3 presents the results obtained from the experiment. Section 4 discusses these results and Section 5 concludes this paper.

\section{Materials and Methods}
\label{sec:methodology}
In this section, we analyzed the data available and evaluated the performance of the ANN method and the configuration of the neural network with reasonable accuracy. 

\subsection{Data}
The data used in this research is based on an open-access dataset collected by \citep {zhang_detection_2019}. This research will use psychometric, saccadic eye movement and DTI data. From an open data set collected and analyzed by \citep{zhang_detection_2019} to address the research question. This data set contains children’s information from 3 to 18 years old. Including subject that are clinically diagnosed with or without FASD.
 
The children are from five different communities in Canada. The data was obtained by a foundation (Kids Brain Health Network, formerly called NeuroDevNet). We compared results with those obtained. Using SVMR and other methods. The data were divided into two groups: a control group and another that includes children who have been diagnosed with FASD. The FASD patient data consists of different neurological condition diagnoses within the FASD group. Such as FAS, pFAS, and ARND. Table \ref{tab:res} summarized the input data information, who many children were diagnosis with FASD and without it from the total data.

\begin{table}[htbp]
    \caption{Data summary from all test}
    \begin{center}
        \begin{tabular}{|c|c|c|c|c|c|}
            \hline
            Test& Number of features (input data)& Total data&FASD&Control \\
            \hline
            \textbf{Psychometric}  &20&129 &58&71  \\
            \hline 
            \textbf{Antisaccade task}&15&174& 68 &106\\
            \hline
            \textbf{Prosaccade task}&18&186&71&115 \\
            \hline
            \textbf{Memory-guide saccade task}&  26&154&61&93 \\
            \hline
            \textbf{DTI}&  48&76&41&35  \\
            \hline
        \end{tabular}
        \label{tab:res}
    \end{center}
\end{table}


The data used describe the performance of children in solve some task. Psychometric tests (NEPSY-II) \citep{paolozza_diffusion_2016} are tests designed regarding the psychosocial, intellectual behavior, memory, and performance of children. Saccadic eye movements is a simple method to infer structural or functional brain deficiencies, present in neurological disorders \citep{green_diffusion_2013}. Ocular behavior such as fixation and saccades are crucial for efficient visual perception. Diffusion tensor image (DTI) is the only non-invasive method for characterizing the microstructural organization of brain tissue \cite{jones_diffusion_2011}. It measures the white matter connectivity in the corpus callosum in great detail with advanced diffusion magnetic resonance (MR) imaging schemes \citep{hagmann_understanding_2006}.

Prenatal alcohol exposure causes brain damage. And the neuropsychological consequences are deep. These deficits in cognitive functions include: difficulties in planning, organization, and attention, consequential learning failures, and memory deficiencies. Some have speech and/or language difficulties, visuospatial functions and spatial memory, that are increased by exposure to prenatal alcohol \citep{glass_academic_2017, green_fetal_2007, mohammad_kcnn2_2020}. Those characteristics could be seen in the task and the evaluation. It is a vital diagnosis based on the abnormalities founded and not only in the traditional facial characteristic \citep{wozniak_microstructural_2009}. The novelty of the Zhang et al. (2019)’s data set is that it includes numeric and image data. This selection observes a deep learning operation in numerical data because most of the studies are based on images.

Prenatal alcohol exposure causes brain damage, and the neuropsychological consequences are deep. These deficits in cognitive functions include difficulties in planning, organization, and attention, consequential learning failures, and memory deficiencies. Some have speech and/or language difficulties, visuospatial functions and spatial memory that are increased by exposure to prenatal alcohol \citep{glass_academic_2017, green_fetal_2007, mohammad_kcnn2_2020}. Those characteristics could be seen in the task and the evaluation is vital diagnosis based on the abnormalities founded and not only in the traditional facial characteristics \citep{wozniak_microstructural_2009}.

\subsection {Data analysis using machine learning} 

Machine learning (ML) is a sub-discipline of artificial intelligence (AI). ML techniques focus on developing algorithms. Capable of learning or adapting to their structure based on observed data. This learning occurs when the optimization function adjusting the weights by calculating the gradient of the loss function. Also known as the cost function, which tells us how good the model is \cite{sajda_machine_2006}. Deep Learning within the ML methods framework, whose theoretical foundation is centered on classical neural networks . Unlike traditional neural networks, these have a group of hidden neurons and layers ($ H_1 ... H_m $) (Fig. \ref{fig:red}) \cite{ravi_deep_2017, jo_deep_2019} . The input data  ($ I_1 ... I_n $) passes through the layers in a non-linear combination of their outputs. Besides, can approximate any arbitrary function through a learning process. To a set of parameters in output projected from input space ($ O_1 ... O_w $) \cite{shukla_deep_2017, deng_new_2013}. The output of each neuron is described as a mathematical formula  \cite {villada_redes_2016}  (See equation \eqref{eq:1}), where $ x_i $ are the weights without apathetic which the weight  $ x_i $ inputs, $ \varphi$ is the neuron activation function and $n$ is the number of neurons connected to the input.

\begin{equation}
    y_{i}=\varphi\left(\sum_{i=0}^{n}w_{ji}x_i \right)
    \label{eq:1}
    \end{equation}
    \begin{figure}[htp]
    \centering
    \include{Red}
    \caption{ANN diagram}
    \label{fig:red}
\end{figure}

The input data are the features of each test. And the output is binary to classify FASD or control. The amount of data must be large enough to provide training examples. From which a large set of parameters can be drawn. Only a large number of parameters give rise to the wealth of class functions that model implicit knowledge \cite{faust_deep_2018}. Unfortunately, there are little data on children with FASD, making the classification difficult. The Keras library of TensorFlow was used in Python to implement the algorithm \citep{ketkar_deep_2017}. We designed various models with dense layer connections, changing the number of neurons in each input layer, hidden layer, and output layer. To compile the model, we used “backpropagation with optimization Adam” and loss “sparse categorical cross-entropy” in the first model. And  “Binary Crossentropy” in another model, which are ideal configurations for classification of categories in the FASD and non-FASD cases.

\subsection {Network configuration}

A study was carried out with numerical data from different tests.To evaluate neural networks’ functioning as an alternative for predicting some neuromental pathology. Then, variations of the type of networks and layers used were made to improve the model’s prediction, along with a feature layer.

\subsubsection {Network configuration for psychometric data}

For the activation of neurons, "Leaky ReLU" (Equation  \eqref{eq2}) was used. It is a function that improves the performance of ReLU, which is the most used in ANN. In the study case, the psychometric data are not in the same range; some are in the range of 1-10 and others of 70-100. Consequently, there is no normalization, and a ReLU activation function could not be the same. Adequate for performance to exceed 80 \% assertiveness in data not observed during training.

\begin{equation}
f(x)= \left\{ \begin{array}{lcc}
             x &   if  & x > 0 \\
             \\ 0.01x &  &otherwise \\
             \end{array}
   \right.
\label{eq2}
\end{equation}

Table \ref{tab1} shows layered neuron configurations in networks. And performance evaluation during training and with the validation data. The psychometric input data is 20. So the neurons in the input layer varied from the same amount of input data to 200. In the output layer, two neurons represent the classification of FASD or not FASD. We made a variation in the number of hidden layers and the number of neurons in them.

The network selection only depended on the {\em accuracy}; we have not carried out {\em cross-validation} of each network to choose the best one. The best performance obtained in the configurations studied was combining 25 neurons in the input layer: a single hidden layer with 20 neurons and two neurons corresponding to the categories that have FASD and no FASD. With this combination, it was possible to obtain a 75.55 \% accuracy with the test data, so it was selected to perform the data analysis. While other combinations showed performance more outstanding than 90\% on training data, performance with unknown or test data was below 65\%.

\begin{table}[htbp]
\caption{Network configuration for performance evaluation with psychometric data.\\ {\scriptsize $^{{a}}$(IL) Input Layer (HL) Hidden Layer. $^{{b}}$(PT) Accuracy in training. (PV) Accuracy in test (unobserved data)}.}
\begin{center}
\begin{tabular}{|c|c|c|c|c|c|}
\hline
\multicolumn{3}{|c|}{\textbf{Number of neurons}}&\multicolumn{2}{|c|}{\textbf{Accuracy}} \\
\hline
\textbf{\textit{IL$^{\mathrm{a}}$}}& \textbf{\textit{1 HL$^{\mathrm{a}}$}}& \textbf{\textit{2 HL$^{\mathrm{a}}$}}& PT$^{\mathrm{b}}$& PV$^{\mathrm{b}}$ \\
\hline
20& 15&-&80.72\% &57.00\%  \\
\hline
25& 15&-&80.72\%& 55.00\%  \\
\hline
25& 20&-&90.24\%&75.55\%  \\
\hline
25& 30& -&85.00\%&65.63\%  \\
\hline
25& 20& 15&88.00\%&65.63\%  \\
\hline
50& 15& -&93.98\% &60.00\%  \\
\hline
100& 50& 25&91.00\%&55.00\%  \\
\hline
200& 15& -&97.59\% &64.00\%  \\
\hline
200& 50& 50&97.00\%&64.00\%  \\
\hline
\end{tabular}
\label{tab1}
\end{center}
\end{table}

\normalsize

Each of the configurations used 1,000 training epochs for performance validation. The evaluation of the model using the training data. We used 75\% of the data for training and 25\% for performance tests. Due to the model result not undergoing improvement, we decided to add a feature layer and increase the number of neurons per layer. Also, we reduce the number of training periods and improving the model by more than 80 \%
\footnote{The algorithms and functions used are available in the GitHub repository \url{https://github.com/vjduarte/ANN_FASD}}.

\subsubsection {Network configuration with addition of eye movement data}

After evaluating the networks’ performance using psychometric data, we could see that the networks did not show a remarkable improvement. This less-than-expected increase is explained partly because the configuration chosen by adding a feature layer to the ANN model.

The proposed network consists of two or more dense layers interconnected for binary classification. A feature layer was added as the first layer to extract information from the input data. This feature layer was allowing the network to learn directly from those characteristics and improve the obtained results. This layer is one of the best ways to propagate the features extracted from all the information that it was found. Under the deep features to the other layers \cite{zhu_dense_2019}.

In the first model, the data set did not include data regarding sex and age to observe the influence of these characteristics. In the prediction of the syndrome, since modifies the data based the age to improve the performance of the models. The results of the evaluation of these models with data not observed during training did not exceed 55\% accuracy. So these characteristics (sex and age) were incorporated in the following models. The general precision improved, exceeding 80\% in all trained models. Depending on the study data, we made different combinations of neurons and layers. Compared to those of the first psychometric study. The models used more neurons in the layers. However, the number of epochs in which the model was trained was lower than the other model.

In the first model, the data set did not include data regarding sex and age to observe the influence of these characteristics. In the prediction of the syndrome, since modifies the data based the age to improve the performance of the models. The results of the evaluation of these models with data not observed during training did not exceed 55\% accuracy. So these characteristics (sex and age) were incorporated in the following models. The general precision improved, exceeding 80\% in all trained models. Depending on the study data, we made different combinations of neurons and layers. Compared to those of the first psychometric study. The models used more neurons in the layers. However, the number of epochs in which the model was trained was lower than the other model. For the Antisaccade movement, Prosaccade used a dense network with two hidden layers of 128 neurons with "ReLu" activation. For Memory-guide saccade, Diffusion tensor imaging (DTI), and Psychometric data, we used four interleaved hidden layers of 64 and 128 neurons with sigmoid, ReLu activation, and Leaky ReLu just for DTI. The models trained for a total of 50 epochs. We used Leaky ReLu just in one case (DTI), which increasing the perfection. In other cases, the results not improved and we discarded to use the optimization function.

\section{Results}

\begin{figure}[htbp] 
    \centering
     \includegraphics[scale=0.5]{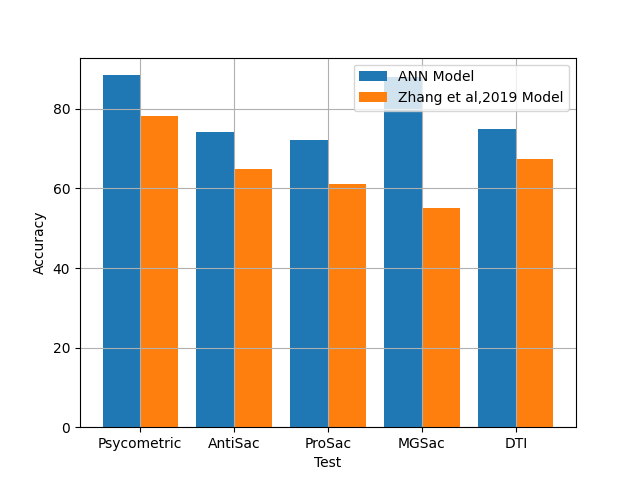}
    \caption{Accuracy model performance comparison by test and by author.}
    \label{fig:Comparacion}
\end{figure}

This research aimed to evaluate the performance of a classification algorithm based on ANN of children with FASD and without the syndrome, using the result of psychometric, DTI, and saccade tests and comparing the accuracy of the model with the SVMR developed by \citeauthor{zhang_detection_2019} et al. (2019). Our results suggest that ANN can make a preliminary diagnosis of pathologies reasonably. When numerical data are available.
Most of the studies carried out to classify FASD patients using machines learning, study brain images or natural language \cite {fang_digital_2006, wozniak_what_2011, suttie_combined_2018}. Only the study conducted by \citeauthor{zhang_detection_2019} et al. (2019) uses numerical data from psychometric and eye movement studies and is freely available. Then, the results obtained with the use of ANN were compared with the results obtained by these authors in their study (Fig. \ref{fig:Comparacion}). The purpose of this study was to evaluate, whether ANN can be used to classify patients with FASD from data obtained non-invasively and that do not require many studies for a preliminary diagnosis.

Results show that our models present a better performance in precision, in all cases than the individual studies carried out by  \citeauthor{zhang_detection_2019}. However, there are key differences that should be considered when it comes to comparing previous results. For instance, \citeauthor{zhang_detection_2019} analyzes the characteristics obtained from each test. And eliminates those that do not contribute to improving the performance of the prediction. In our study, all the features obtained from each study are left, and using a feature layer allows the algorithm to distinguish those characteristics. Which contribute to improving pathology’s prediction accuracy. In turn, we did not specify which will be the training and test group. But instead it is left with a random division (80\% training data, 20\% test data). Comparing the structure of the data of each test, we could note that some tests had significantly more samples of control patients than FASD patients. Thus, the results showed a high percentage of accuracy to predict control but not to predict FASD. We eliminated samples from control patients without any specific discrimination. And we evaluate the models to avoid imbalance issues.

In addition, we realized that by eliminating sex and age from the studies, the prediction decreased dramatically. So any study must consider these characteristics in its application. With the use of ANN, we noted that on average, ANN works 14.22\% better than the results obtained by with a standard deviation of 10.58.

\subsection{Pyscometric data classification}

The data used for testing included psychometric tests. Associated with analyzing social behavior, memory activities, language delay and all altered behavioral factors in FASD children. Also, the data set included other developmental diseases, making the classification process more difficult. This paper used two models of dense networks using this type of data. The first model archived an accuracy of 75.55\%. However, the model did not show significant improvements, despite the variation in the layers and neurons in the layers. In the second model, a feature layer was added, achieving an accuracy of 88.46\%.

\begin{figure}[H]
     \centering
     \begin{subfigure}[b]{0.36\textwidth}
         \centering
         \includegraphics[width=\textwidth]{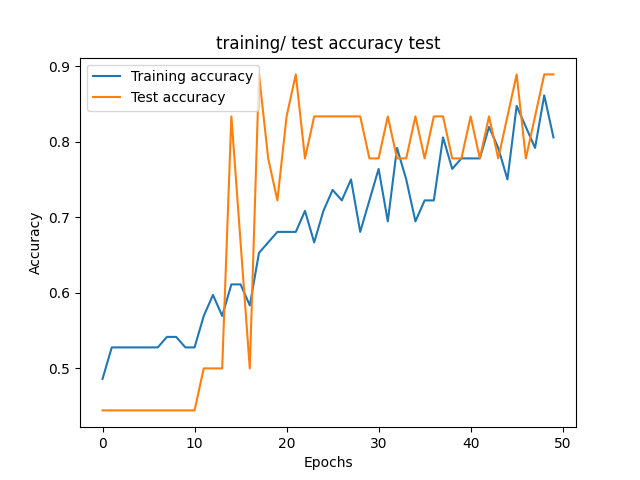}
         \caption{$Accuracy$}
         \label{fig:PsycoAcc}
     \end{subfigure}
     \hfill
     \begin{subfigure}[b]{0.36\textwidth}
         \centering
         \includegraphics[width=\textwidth]{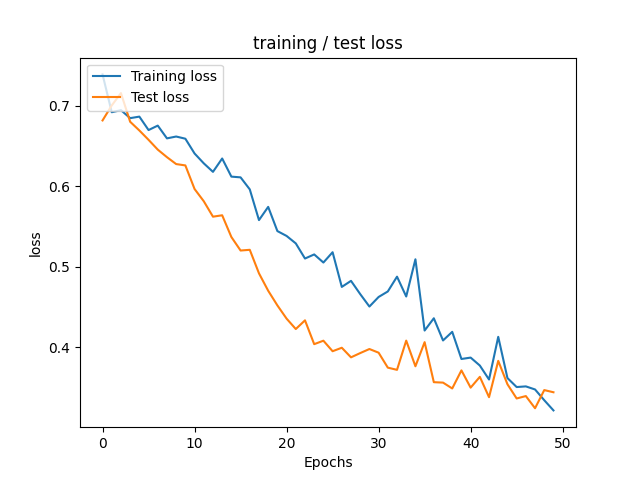}
         \caption{$Loss$}
         \label{fig:PsyLoss}
     \end{subfigure}
       \caption{Precision and Loss using feature dense layer on psychometric data} 
\end{figure}

Once the network configuration was chosen, training and testing or validation behavior are shown in the model precision and loss functions. The loss function with a high result indicates that the neural network has a poor performance and a low result, that it is doing a good job. Fig. \ref{fig:PsycoAcc} shows the accuracy for each of the data set. In terms of the number of that it was aspects found and related to the number of individuals evaluated. We attempted to find a model with no significant difference, between the labeled data and the prediction. Nonetheless, with the implemented configuration, the training data has increased accuracy. In addition, the validation data are also increasing accuracy.

Fig. \ref{fig:PsycoAcc} shows the loss function on the psychometric data. In this case, the loss function has a decreasing curve in both training and testing. This result suggests that the network may have issues classifying, some data not used during training but not significantly.

The confusion matrix (Fig. \ref{fig:PFL}) shows the performance of the model with test data. The model has a good performance in classifying children with FASD since the correctness is 38.46\% in positive hits (FASD having FASD), and it has 7.69\% false positives (a control having FASD). It should be noted that the total percentage of children diagnosed with FASD is 42.31\%. So, the model allows classifying FASD in most cases. In classifying control patients, the model classifies 51.0\% of control patients who are really in control. And fails in 3.85\% to classify patients with FASD as controls.

\begin{figure}[hbtp!]
    \centering
\input{./Figures/MCPsyFD.tex}
    \caption{Classification Confusion matrix using psychometric test}
    \label{fig:PFL}
\end{figure}

\subsection{Antisaccad data classification}
The results from the Antisaccade tasks correspond to measurements of successes between the number of tests achieved by each individual, where the failures in the automatic saccade inhibition with a peripheral objective are measured. In this case, initially, 173 records represented 61\% control patients and only 39\% FASD patients, so we eliminated some control patient records, leaving fewer observations but whose representation of both groups was 50\% for each. 

\begin{figure}[htbp]
    \centering
\input{./Figures/MCAntiSac.tex}
    \caption{Classification Confusion matrix using Antisaccade task}
    \label{fig:CMAntSac}
\end{figure}

The confusion matrix (Fig. \ref{fig:CMAntSac}) results show a high rate of prediction of classifying FASD compared to the amount of data evaluated. However, some of the classification characteristics tend to confuse control children with the possibility of having FASD. These results make it necessary to evaluate the degree of FASD these children have and the reason for this classification. 

\subsection{Prosaccade task data classification}

Prosaccade tests measure reaction time, performance accuracy, viability, and parameters which corresponding to the main sequence. For the classification of FASD patients with these data from the 186 records, only 38\% correspond to patients diagnosed with FASD. Accordingly, only a percentage was used that allowed for equality between control and FASD patients. They are achieving an accuracy of 72.41

The confusion matrix (Fig. \ref{fig:CMProSac}) for this case shows that the model predicts FASD patients quite well, but only in 50\% of the cases control patients being controls. This confusion matrix allows us to observe that we just made some adjustments to improve the exactness to predict control patients or verify the eliminated data. We should better understand the present differentiating characteristics from control patients to adequately identified during training reflected in the test data.
\begin{figure}[htbp]
    \centering
\input{./Figures/MCProSac.tex}
    \caption{Classification Confusion matrix using Prosaccade task }
    \label{fig:CMProSac}
\end{figure}

\subsection {Memory-guided saccade task classification results}

These tasks are related to following the objective order; When there are errors, the subject cannot follow the order that initiates the saccade by the second objective more than on the first. In these tasks, time and errors are measured \cite{zhang_detection_2019}. We took 61 records by classification, achieving 88\% in test data, one of the highest in the study.

\begin{figure}[htbp]
    \centering
\input{./Figures/MCMGSac.tex}
    \caption{Classification Confusion matrix using Memory-guide saccadic task}
    \label{fig:CMMG}
\end{figure}

The confusion matrix (Fig. [fig:CMMG]) for this model shows a balanced measurement of FASD and control patients greater than 80\% per case. And in the case of FASD, very few mistakes. It is important to note that the studies that achieved the best results have an element of the individual’s memory. This result makes us think, that we must pay more attention to this type of test, including memory issues.

\subsection {Diffusion tensor imaging (DTI) data classification results}

This study measures the connectivity of the white matter in the corpus callosum. It comes from a structural MRI, so it does not involve tests . This study has 76.54\% records classified as FASD and 46\% control. Having little data means that this study did not achieve a good accuracy (less than 50\%). The Leaky-ReLU function was added in the intermediate layers to improve. Managing to rise to 75\% in a combination of 4 hidden layers of 64 and 128 neurons, a feature layer, and the training time was 100 epochs compared to the other models with 50 epochs training.

\begin{figure}[htbp]
    \centering
\input{./Figures/MCDTI.tex}
    \caption{Classification Confusion matrix using DTI}
    \label{fig:CMDTI}
\end{figure}

The confusion matrix (Fig. \ref{fig:CMDTI}) shows that the percentage of failures to classify FASD and control is the same. And concerning the percentage of samples of each type, observing a good performance.

\section{Discussion}
This paper evaluated the usage of computational algorithms. Based on ANN to classify children with FASD and, this is why can be used for medical diagnosis. This paper estimated the prediction model using psychometric data, DTI, and saccadic eye movement. Most of the current research that implements ANN to predict FASD is focused on the study of images, the novelty of this paper is that used numerical data to estimate the model. Although Zhang et al. (2019) \citep{zhang_detection_2019} collect and use the same data, our results show a better performance using ANN instead of SVMR or other methods.

Zhang et al. (2019) \citep{zhang_detection_2019} utilized methods, especially SVMR, require fewer parameters to optimize its performance, reducing the possibility of over-tuning training data, and increasing actual performance. According to the training data results -unlike ANN- the study of different configurations can change its operation and performance in general. The increase in training data allows ANN to improve its performance. Without changing parameters. Nevertheless a new training improves accuracy, unlike SVMR. SVMR is faster and more stable than other ANN. So the implementation of one or the other model depends on the change in data or other characteristics rather than on the accuracy of each one.
Consequently, this paper explored the usage of numerical data, instead of images for diagnosing FASD. First, for psychometric data, a basic network configuration was obtained with correctness with test data of 75.5\% with a Leaky-ReLU activation function. Which is recommended for data without normalization, as in this case. It is expected that the models using machine learning will predict the output data in at least 90\% of the cases. However, the model and data used, it was not achieved, remaining below expectations, by a 15\%. Subsequently, a change was made in the configuration of the networks. And a feature layer was added. This layer allows identifying the essential characteristics. This significantly contribute to the classification, achieving 88\% accuracy. So with this configuration; it was close to the desired prediction value of at least 90\%, being a preliminary diagnosis and not achieving a reasonable prediction could have consequences. This is why for the other studies carried out, we used a network configuration with a feature layer to observe the operation of these networks. With other data of numerical origin related to FASD, achieving in all cases an accuracy greater than 70\% above from achieved by .
The confusion matrices obtained demonstrate that the model predicts at least 71\% of patients diagnosed with FASD. From the around 50\% that this group represents. However, the prediction of control patients has over 77 \% of patients who have not been diagnosed with FASD (except Prosaccadic task). These matrices suggest that the model must be recalibrated or overlearned from the data of patients with FASD, or else some control patients may have FASD but have not been diagnosed. These results cannot be proven. So according to the data, the use of ANN as a diagnostic tool must be improved before its use can be suggested.
One caveat is the occurrence of the false-positive. That is to diagnose a person with FASD that does not have it since the data between FASD and control patients can be similar. It is essential to mention that some patients are not correctly diagnosed with FASD or have some cognitive problems similar to FASD. Which can also lead to many false negatives. This result is because collecting data on alcohol consumption during pregnancy is not accessible. Yet it is requires the mother’s declaration about the consumption. Likewise, over time the early samples of FASD decrease and are dependent on the time of exposure to alcohol. And also the phase of gestation during which the ingestion occurred, and the amount of it. The neuropsychological and the developmental deficits which some patients present (between groups diagnosed with FASD and other pathologies) do not show significant differences, disabilities, or behavioral problems. Therefore, the diagnosis of FASD cannot be made lightly and should be considered other results such as magnetic resonance imaging, eye movement, among others.

\section{Conclusions}

In this research, ANN algorithms were developed to classify: psychometric data, DTI, and Saccadic eye movement of children/young people diagnosed with FASD. These data were trained in different network configurations. Measuring their accuracy in training and test data (data not used during training). We achieve an exactness of over 70\% on all models with the network configuration selected. Its performance in training was over 90\%. The selected model was studied based on the loss and accuracy functions. Evaluating its performance in predicting FASD, control, and false positives or false negatives. Currently, there is an underestimation of people with FASD. This syndrome affects a significant percentage of the world population. However, its diagnosis requires the certainty of consumption during embryonic development. It could be undiagnosed in our control population due to the clinical similarities with other neuro-development diseases.
In conclusion, we suggest that ANN algorithms improve their performance with a suitable network configuration -increase in the number of neurons, hidden layers, change in optimization functions, and feature layer- and the activation algorithm used to incorporate new data into training. The combination with other data groups even allows the increase of its performance, which is why ANN can be a good alternative in machine learning algorithms. However, a deeper study of the different network configurations must be undertaken to improve the prediction.

\bibliography{ref}
\bibliographystyle{apalike}
\end{document}

%% file: Figures/MCPsyFD.tex
\newcommand{\daywidth}{2 cm}
\begin{tikzpicture}[x=\daywidth, y=-1cm, node distance=0 cm,outer sep = 0pt]
\tikzstyle{day}=[draw, rectangle,  minimum height=1cm, minimum width=\daywidth, fill=yellow!20,anchor=south west]
\tikzstyle{hour}=[draw, rectangle, minimum height=1 cm, minimum width=1.5 cm, fill=yellow!30,anchor=north east]

\tikzstyle{hours}=[rectangle,draw, minimum width=\daywidth, anchor=north west,text centered,text width=5 em]
\tikzstyle{1hour}=[hours,minimum height=1cm]
\tikzstyle{2hours}=[hours,minimum height=2cm]
\tikzstyle{3hours}=[hours,minimum height=3cm]
\tikzstyle{Planche}=[1hour,fill=white]
\tikzstyle{PP}=[1hour,fill=blue!10]
\tikzstyle{FN}=[1hour,fill=green!20]
\node[day] (lundi) at (1,8) {FASD};
\node[day] (mardi) [right = of lundi] {Control};
\node[hour] (8-9) at (1,8) {FASD};
\node[hour] (9-10) [below = of 8-9] {Control};
\node[PP] at (1,8) {38.46\%}; \node[FN] at (1,9) {3.85\%}; \node[Planche] at (1,10) {42.31\%};
\node[FN] at (2,8) {7.69\%}; \node[PP] at (2,9) {50.00\%}; \node[Planche] at (2,10) {57.69\%};
\node[Planche] at (3,8) {46.15\%}; \node[Planche] at (3,9) {53.85\%};

\end{tikzpicture}

%% file: Figures/MCAntiSac.tex
\newcommand{\daywidth}{2 cm}
\begin{tikzpicture}[x=\daywidth, y=-1cm, node distance=0 cm,outer sep = 0pt]
\tikzstyle{day}=[draw, rectangle,  minimum height=1cm, minimum width=\daywidth, fill=yellow!20,anchor=south west]
\tikzstyle{hour}=[draw, rectangle, minimum height=1 cm, minimum width=1.5 cm, fill=yellow!30,anchor=north east]

\tikzstyle{hours}=[rectangle,draw, minimum width=\daywidth, anchor=north west,text centered,text width=5 em]
\tikzstyle{1hour}=[hours,minimum height=1cm]
\tikzstyle{2hours}=[hours,minimum height=2cm]
\tikzstyle{3hours}=[hours,minimum height=3cm]
\tikzstyle{Planche}=[1hour,fill=white]
\tikzstyle{PP}=[1hour,fill=blue!10]
\tikzstyle{FN}=[1hour,fill=green!20]
\node[day] (lundi) at (1,8) {FASD};
\node[day] (mardi) [right = of lundi] {Control};
\node[hour] (8-9) at (1,8) {FASD};
\node[hour] (9-10) [below = of 8-9] {Control};
\node[PP] at (1,8) {22.22\%}; \node[FN] at (1,9) {0.0\%}; \node[Planche] at (1,10) {22.22\%};
\node[FN] at (2,8) {25.93\%}; \node[PP] at (2,9) {51.85\%}; \node[Planche] at (2,10) {77.78\%};
\node[Planche] at (3,8) {48.15\%}; \node[Planche] at (3,9) {51.85\%};

\end{tikzpicture}

%% file: Figures/MCProSac.tex
\newcommand{\daywidth}{2 cm}
\begin{tikzpicture}[x=\daywidth, y=-1cm, node distance=0 cm,outer sep = 0pt]
\tikzstyle{day}=[draw, rectangle,  minimum height=1cm, minimum width=\daywidth, fill=yellow!20,anchor=south west]
\tikzstyle{hour}=[draw, rectangle, minimum height=1 cm, minimum width=1.5 cm, fill=yellow!30,anchor=north east]

\tikzstyle{hours}=[rectangle,draw, minimum width=\daywidth, anchor=north west,text centered,text width=5 em]
\tikzstyle{1hour}=[hours,minimum height=1cm]
\tikzstyle{2hours}=[hours,minimum height=2cm]
\tikzstyle{3hours}=[hours,minimum height=3cm]
\tikzstyle{Planche}=[1hour,fill=white]
\tikzstyle{PP}=[1hour,fill=blue!10]
\tikzstyle{FN}=[1hour,fill=green!20]
\node[day] (lundi) at (1,8) {FASD};
\node[day] (mardi) [right = of lundi] {Control};
\node[hour] (8-9) at (1,8) {FASD};
\node[hour] (9-10) [below = of 8-9] {Control};
\node[PP] at (1,8) {51.72\%}; \node[FN] at (1,9) {6.9\%}; \node[Planche] at (1,10) {58.62\%};
\node[FN] at (2,8) {20.69\%}; \node[PP] at (2,9) {20.69\%}; \node[Planche] at (2,10) {41.38\%};
\node[Planche] at (3,8) {72.41\%}; \node[Planche] at (3,9) {27.59\%};

\end{tikzpicture}

%% file: Figures/MCMGSac.tex
\newcommand{\daywidth}{2 cm}
\begin{tikzpicture}[x=\daywidth, y=-1cm, node distance=0 cm,outer sep = 0pt]
\tikzstyle{day}=[draw, rectangle,  minimum height=1cm, minimum width=\daywidth, fill=yellow!20,anchor=south west]
\tikzstyle{hour}=[draw, rectangle, minimum height=1 cm, minimum width=1.5 cm, fill=yellow!30,anchor=north east]

\tikzstyle{hours}=[rectangle,draw, minimum width=\daywidth, anchor=north west,text centered,text width=5 em]
\tikzstyle{1hour}=[hours,minimum height=1cm]
\tikzstyle{2hours}=[hours,minimum height=2cm]
\tikzstyle{3hours}=[hours,minimum height=3cm]
\tikzstyle{Planche}=[1hour,fill=white]
\tikzstyle{PP}=[1hour,fill=blue!10]
\tikzstyle{FN}=[1hour,fill=green!20]
\node[day] (lundi) at (1,8) {FASD};
\node[day] (mardi) [right = of lundi] {Control};
\node[hour] (8-9) at (1,8) {FASD};
\node[hour] (9-10) [below = of 8-9] {Control};
\node[PP] at (1,8) {36.0\%}; \node[FN] at (1,9) {8.0\%}; \node[Planche] at (1,10) {44.0\%};
\node[FN] at (2,8) {4.0\%}; \node[PP] at (2,9) {52.0\%}; \node[Planche] at (2,10) {56.0\%};
\node[Planche] at (3,8) {40.0\%}; \node[Planche] at (3,9) {60.0\%};

\end{tikzpicture}

%% file: Figures/MCDTI.tex
\newcommand{\daywidth}{2 cm}
\begin{tikzpicture}[x=\daywidth, y=-1cm, node distance=0 cm,outer sep = 0pt]
\tikzstyle{day}=[draw, rectangle,  minimum height=1cm, minimum width=\daywidth, fill=yellow!20,anchor=south west]
\tikzstyle{hour}=[draw, rectangle, minimum height=1 cm, minimum width=1.5 cm, fill=yellow!30,anchor=north east]

\tikzstyle{hours}=[rectangle,draw, minimum width=\daywidth, anchor=north west,text centered,text width=5 em]
\tikzstyle{1hour}=[hours,minimum height=1cm]
\tikzstyle{2hours}=[hours,minimum height=2cm]
\tikzstyle{3hours}=[hours,minimum height=3cm]
\tikzstyle{Planche}=[1hour,fill=white]
\tikzstyle{PP}=[1hour,fill=blue!10]
\tikzstyle{FN}=[1hour,fill=green!20]
\node[day] (lundi) at (1,8) {FASD};
\node[day] (mardi) [right = of lundi] {Control};
\node[hour] (8-9) at (1,8) {FASD};
\node[hour] (9-10) [below = of 8-9] {Control};
\node[PP] at (1,8) {31.25\%}; \node[FN] at (1,9) {12.5\%}; \node[Planche] at (1,10) {43.75\%};
\node[FN] at (2,8) {12.5\%}; \node[PP] at (2,9) {43.75\%}; \node[Planche] at (2,10) {56.25\%};
\node[Planche] at (3,8) {43.75\%}; \node[Planche] at (3,9) {56.25\%};

\end{tikzpicture}

%% file: template.bbl
\begin{thebibliography}{}

\bibitem[Al-Shayea, 2011]{al2011artificial}
Al-Shayea, Q.~K. (2011).
\newblock Artificial neural networks in medical diagnosis.
\newblock {\em International Journal of Computer Science Issues},
  8(2):150--154.

\bibitem[Chudley, 2018]{chudley_diagnosis_2018}
Chudley, A.~E. (2018).
\newblock Diagnosis of fetal alcohol spectrum disorder: current practices and
  future considerations.
\newblock {\em Biochem Cell Biol}, 96(2):231--236.

\bibitem[Cook et~al., 2016]{cook_fetal_2016}
Cook, J.~L., Green, C.~R., Lilley, C.~M., Anderson, S.~M., Baldwin, M.~E.,
  Chudley, A.~E., Conry, J.~L., LeBlanc, N., Loock, C.~A., Lutke, J., Mallon,
  B.~F., McFarlane, A.~A., Temple, V.~K., and Rosales, T. (2016).
\newblock Fetal alcohol spectrum disorder: a guideline for diagnosis across the
  lifespan.
\newblock {\em CMAJ}, 188(3):191--197.

\bibitem[Deng et~al., 2013]{deng_new_2013}
Deng, L., Hinton, G., and Kingsbury, B. (2013).
\newblock New types of deep neural network learning for speech recognition and
  related applications: an overview.
\newblock In {\em 2013 {IEEE} {International} {Conference} on {Acoustics},
  {Speech} and {Signal} {Processing}}, pages 8599--8603, Vancouver, Canada.

\bibitem[El~Naqa and Murphy, 2015]{el2015machine}
El~Naqa, I. and Murphy, M.~J. (2015).
\newblock What is machine learning?
\newblock In {\em machine learning in radiation oncology}, pages 3--11.
  Springer.

\bibitem[Fang et~al., 2006]{fang_digital_2006}
Fang, J., Fang, S., Huang, J., and Tuceryan, M. (2006).
\newblock Digital geometry image analysis for medical diagnosis.
\newblock In {\em Proceedings of the 2006 {ACM Symposium on Applied
  Computing}}, {SAC} '06, pages 217--221, Dijon, France. ACM.

\bibitem[Fang et~al., 2009]{fang_facial_2009}
Fang, S., Liu, Y., Huang, J., Vinci-Booher, S., Anthony, B., and Zhou, F.
  (2009).
\newblock Facial {Image} {Classification} of {Mouse} {Embryos} for the {Animal}
  {Model} {Study} of {Fetal} {Alcohol} {Syndrome}.
\newblock {\em Proc. Symp. Appl. Comput.}, 2009:852--856.

\bibitem[Faust et~al., 2018]{faust_deep_2018}
Faust, O., Hagiwara, Y., Hong, T.~J., Lih, O.~S., and Acharya, U.~R. (2018).
\newblock Deep learning for healthcare applications based on physiological
  signals: {A} review.
\newblock {\em Computer Methods and Programs in Biomedicine}, 161:1--13.

\bibitem[Glass et~al., 2017]{glass_academic_2017}
Glass, L., Moore, E.~M., Akshoomoff, N., Jones, K.~L., Riley, E.~P., and
  Mattson, S.~N. (2017).
\newblock Academic {Difficulties} in {Children} with {Prenatal} {Alcohol}
  {Exposure}: {Presence}, {Profile}, and {Neural} {Correlates}.
\newblock {\em Alcohol Clin Exp Res}, 41(5):1024--1034.

\bibitem[Green et~al., 2013]{green_diffusion_2013}
Green, C.~R., Lebel, C., Rasmussen, C., Beaulieu, C., and Reynolds, J.~N.
  (2013).
\newblock Diffusion {Tensor} {Imaging} {Correlates} of {Saccadic} {Reaction}
  {Time} in {Children} with {Fetal} {Alcohol} {Spectrum} {Disorder}.
\newblock {\em Alcoholism: Clinical and Experimental Research},
  37(9):1499--1507.

\bibitem[Green, 2007]{green_fetal_2007}
Green, J.~H. (2007).
\newblock Fetal {Alcohol} {Spectrum} {Disorders}: {Understanding} the {Effects}
  of {Prenatal} {Alcohol} {Exposure} and {Supporting} {Students}.
\newblock {\em Journal of School Health}, 77(3):103--108.

\bibitem[Hagmann et~al., 2006]{hagmann_understanding_2006}
Hagmann, P., Jonasson, L., Maeder, P., Thiran, J.-P., Wedeen, V.~J., and Meuli,
  R. (2006).
\newblock Understanding {Diffusion} {MR} {Imaging} {Techniques}: {From}
  {Scalar} {Diffusion}-weighted {Imaging} to {Diffusion} {Tensor} {Imaging} and
  {Beyond}.
\newblock {\em RadioGraphics}, 26(suppl\_1):S205--S223.

\bibitem[Huang et~al., 2005]{huang_using_2005}
Huang, J., Jain, A., Fang, S., and Riley, E. (2005).
\newblock Using facial images to diagnose fetal alcohol syndrome ({FAS}).
\newblock In {\em International {Conference} on {Information} {Technology}:
  {Coding} and {Computing} ({ITCC}'05) - {Volume} {II}}, pages 66--71 Vol. 2,
  Las Vegas, NV, USA. IEEE.

\bibitem[Jo et~al., 2019]{jo_deep_2019}
Jo, T., Nho, K., and Saykin, A.~J. (2019).
\newblock Deep {Learning} in {Alzheimer}'s {Disease}: {Diagnostic}
  {Classification} and {Prognostic} {Prediction} {Using} {Neuroimaging} {Data}.
\newblock {\em Front. Aging Neurosci.}, 11.
\newblock Publisher: Frontiers.

\bibitem[Jones and Leemans, 2011]{jones_diffusion_2011}
Jones, D.~K. and Leemans, A. (2011).
\newblock Diffusion {Tensor} {Imaging}.
\newblock In Modo, M. and Bulte, J.~W., editors, {\em Magnetic {Resonance}
  {Neuroimaging}: {Methods} and {Protocols}}, Methods in {Molecular} {Biology},
  pages 127--144. Humana Press, Totowa, NJ.

\bibitem[Ketkar, 2017]{ketkar_deep_2017}
Ketkar, N. (2017).
\newblock {\em Deep {Learning} with {Python}: {A} {Hands}-on {Introduction}}.
\newblock Apress.

\bibitem[Kim et~al., 2011]{kim_pre-operative_2011}
Kim, S.~Y., Moon, S.~K., Jung, D.~C., Hwang, S.~I., Sung, C.~K., Cho, J.~Y.,
  Kim, S.~H., Lee, J., and Lee, H.~J. (2011).
\newblock Pre-operative prediction of advanced prostatic cancer using clinical
  decision support systems: accuracy comparison between support vector machine
  and artificial neural network.
\newblock {\em Korean J Radiol}, 12(5):588--594.

\bibitem[Li et~al., 2019]{li_deep_2019}
Li, Y., Huang, C., Ding, L., Li, Z., Pan, Y., and Gao, X. (2019).
\newblock Deep learning in bioinformatics: {Introduction}, application, and
  perspective in the big data era.
\newblock {\em Methods}, 166:4--21.

\bibitem[Little and Beaulieu, 2020]{little_multivariate_2020}
Little, G. and Beaulieu, C. (2020).
\newblock Multivariate models of brain volume for identification of children
  and adolescents with fetal alcohol spectrum disorder.
\newblock {\em Human Brain Mapping}, 41(5):1181--1194.

\bibitem[Lundervold and Lundervold, 2019]{lundervold_overview_2019}
Lundervold, A.~S. and Lundervold, A. (2019).
\newblock An overview of deep learning in medical imaging focusing on {MRI}.
\newblock {\em Zeitschrift für Medizinische Physik}, 29(2):102--127.

\bibitem[May et~al., 2018]{may_prevalence_2018}
May, P.~A., Chambers, C.~D., Kalberg, W.~O., Zellner, J., Feldman, H., Buckley,
  D., Kopald, D., Hasken, J.~M., Xu, R., Honerkamp-Smith, G., Taras, H.,
  Manning, M.~A., Robinson, L.~K., Adam, M.~P., Abdul-Rahman, O., Vaux, K.,
  Jewett, T., Elliott, A.~J., Kable, J.~A., Akshoomoff, N., Falk, D., Arroyo,
  J.~A., Hereld, D., Riley, E.~P., Charness, M.~E., Coles, C.~D., Warren,
  K.~R., Jones, K.~L., and Hoyme, H.~E. (2018).
\newblock Prevalence of {Fetal} {Alcohol} {Spectrum} {Disorders} in 4 {US}
  {Communities}.
\newblock {\em JAMA}, 319(5):474.

\bibitem[McCallum and Holland, 2018]{mccallum2018drink}
McCallum, K. and Holland, K. (2018).
\newblock ‘to drink or not to drink’: media framing of evidence and debate
  about alcohol consumption in pregnancy.
\newblock {\em Critical Public Health}, 28(4):412--423.

\bibitem[Mohammad et~al., 2020]{mohammad_kcnn2_2020}
Mohammad, S., Page, S.~J., Wang, L., Ishii, S., Li, P., Sasaki, T., Basha, A.,
  Salzberg, A., Quezado, Z., Imamura, F., Nishi, H., Isaka, K., Corbin, J.~G.,
  Liu, J.~S., Kawasawa, Y.~I., Torii, M., and Hashimoto-Torii, K. (2020).
\newblock Kcnn2 blockade reverses learning deficits in a mouse model of fetal
  alcohol spectrum disorders.
\newblock {\em Nature Neuroscience}, 23(4):533--543.

\bibitem[O'Leary and Bower, 2012]{o2012guidelines}
O'Leary, C. and Bower, C. (2012).
\newblock Guidelines for pregnancy: what's an acceptable risk, and how is the
  evidence (finally) shaping up?
\newblock {\em Drug and alcohol review}, 31(2):170--183.

\bibitem[Paolozza et~al., 2013]{paolozza_altered_2013}
Paolozza, A., Titman, R., Brien, D., Munoz, D.~P., and Reynolds, J.~N. (2013).
\newblock Altered {Accuracy} of {Saccadic} {Eye} {Movements} in {Children} with
  {Fetal} {Alcohol} {Spectrum} {Disorder}.
\newblock {\em Alcoholism: Clinical and Experimental Research},
  37(9):1491--1498.

\bibitem[Paolozza et~al., 2016]{paolozza_diffusion_2016}
Paolozza, A., Treit, S., Beaulieu, C., and Reynolds, J.~N. (2016).
\newblock Diffusion tensor imaging of white matter and correlates to eye
  movement control and psychometric testing in children with prenatal alcohol
  exposure.
\newblock {\em Hum Brain Mapp}, 38(1):444--456.

\bibitem[Popova et~al., 2018]{popova2018world}
Popova, S., Lange, S., Chudley, A.~E., Reynolds, J.~N., Rehm, J., May, P., and
  Riley, E. (2018).
\newblock World health organization international study on the prevalence of
  fetal alcohol spectrum disorder (fasd).
\newblock {\em Cent. Addit. Ment. Heal}.

\bibitem[Popova et~al., 2017]{popova_estimation_2017}
Popova, S., Lange, S., Probst, C., Gmel, G., and Rehm, J. (2017).
\newblock Estimation of national, regional, and global prevalence of alcohol
  use during pregnancy and fetal alcohol syndrome: a systematic review and
  meta-analysis.
\newblock {\em The Lancet Global Health}, 5(3):e290--e299.

\bibitem[Ravi et~al., 2017]{ravi_deep_2017}
Ravi, D., Wong, C., Deligianni, F., Berthelot, M., Andreu-Perez, J., Lo, B.,
  and Yang, G.-Z. (2017).
\newblock Deep {Learning} for {Health} {Informatics}.
\newblock {\em IEEE J Biomed Health Inform}, 21(1):4--21.

\bibitem[Rodriguez et~al., 2021]{rodriguez2021detection}
Rodriguez, C.~I., Vergara, V.~M., Davies, S., Calhoun, V.~D., Savage, D.~D.,
  and Hamilton, D.~A. (2021).
\newblock Detection of prenatal alcohol exposure using machine learning
  classification of resting-state functional network connectivity data.
\newblock {\em Alcohol}, 93:25--34.

\bibitem[Sajda, 2006]{sajda_machine_2006}
Sajda, P. (2006).
\newblock Machine {Learning} for {Detection} and {Diagnosis} of {Disease}.
\newblock {\em Annual Review of Biomedical Engineering}, 8(1):537--565.

\bibitem[Shukla et~al., 2017]{shukla_deep_2017}
Shukla, P., Gupta, T., Saini, A., Singh, P., and Balasubramanian, R. (2017).
\newblock A {Deep} {Learning} {Frame}-{Work} for {Recognizing} {Developmental}
  {Disorders}.
\newblock In {\em 2017 {IEEE} {Winter} {Conference} on {Applications} of
  {Computer} {Vision} ({WACV})}, pages 705--714.

\bibitem[Sun et~al., 2017]{sun_revisiting_2017}
Sun, C., Shrivastava, A., Singh, S., and Gupta, A. (2017).
\newblock Revisiting {Unreasonable} {Effectiveness} of {Data} in {Deep}
  {Learning} {Era}.
\newblock pages 843--852.

\bibitem[Suttie et~al., 2018]{suttie_combined_2018}
Suttie, M., Wozniak, J.~R., Parnell, S.~E., Wetherill, L., Mattson, S.~N.,
  Sowell, E.~R., Kan, E., Riley, E.~P., Jones, K.~L., Coles, C., Foroud, T.,
  and Hammond, P. (2018).
\newblock Combined {Face}–{Brain} {Morphology} and {Associated}
  {Neurocognitive} {Correlates} in {Fetal} {Alcohol} {Spectrum} {Disorders}.
\newblock {\em Alcoholism: Clinical and Experimental Research},
  42(9):1769--1782.

\bibitem[Villada et~al., 2016]{villada_redes_2016}
Villada, F., Mu\~noz, N., and Garc\'ia-Quintero, E. (2016).
\newblock Redes {Neuronales} {Artificiales} aplicadas a la {Predicci\'on} del
  {Precio} del {Oro}.
\newblock {\em Informaci\'on tecnol\'ogica}, 27(5):143--150.

\bibitem[Wozniak and Muetzel, 2011]{wozniak_what_2011}
Wozniak, J.~R. and Muetzel, R.~L. (2011).
\newblock What does diffusion tensor imaging reveal about the brain and
  cognition in fetal alcohol spectrum disorders?
\newblock {\em Neuropsychol Rev}, 21(2):133--147.

\bibitem[Wozniak et~al., 2009]{wozniak_microstructural_2009}
Wozniak, J.~R., Muetzel, R.~L., Mueller, B.~A., McGee, C.~L., Freerks, M.~A.,
  Ward, E.~E., Nelson, M.~L., Chang, P.-N., and Lim, K.~O. (2009).
\newblock Microstructural {Corpus} {Callosum} {Anomalies} in {Children} {With}
  {Prenatal} {Alcohol} {Exposure}: {An} {Extension} of {Previous} {Diffusion}
  {Tensor} {Imaging} {Findings}.
\newblock {\em Alcoholism: Clinical and Experimental Research},
  33(10):1825--1835.

\bibitem[Zhang et~al., 2019]{zhang_detection_2019}
Zhang, C., Paolozza, A., Tseng, P.-H., Reynolds, J.~N., Munoz, D.~P., and Itti,
  L. (2019).
\newblock Detection of {Children}/{Youth} {With} {Fetal} {Alcohol} {Spectrum}
  {Disorder} {Through} {Eye} {Movement}, {Psychometric}, and {Neuroimaging}
  {Data}.
\newblock {\em Front. Neurol.}, 10.

\bibitem[Zhu et~al., 2019]{zhu_dense_2019}
Zhu, Y., Li, C., Luo, B., Tang, J., and Wang, X. (2019).
\newblock Dense {Feature} {Aggregation} and {Pruning} for {RGBT} {Tracking}.
\newblock In {\em Proceedings of the 27th {ACM} {International} {Conference} on
  {Multimedia}}, {MM} '19, pages 465--472, New York, NY, USA. Association for
  Computing Machinery.

\end{thebibliography}
